%% file: main.tex
\documentclass[letterpaper, 10 pt, conference]{ieeeconf}  

\IEEEoverridecommandlockouts                              

\usepackage{graphicx} 
\usepackage{amsmath} 
\usepackage{amssymb}  
\usepackage[inkscapelatex=false]{svg}

\usepackage{cite}
\usepackage[colorlinks=false,urlcolor=blue]{hyperref}

\usepackage[caption=false]{subfig}
\usepackage{tabularx}
\usepackage{wasysym}
\usepackage{booktabs}

\usepackage{soul}
\usepackage[dvipsnames]{xcolor}

\usepackage{needspace}

\title{\LARGE \bf
Scalable Open-Source Visuotactile Sensor for 6-Axis Contact Wrench Estimation in Tensegrity Robots
}

\author{Wenzhe Tong$^*$, Jonathan Mi$^*$, Xili Yi, Nima Fazeli, and Xiaonan Huang
    \thanks{$^*$ Equal contributions.}
    \thanks{
    This work was supported in part by the startup fund from the Robotics Department at the University of Michigan and in part by the National Science Foundation through the Graduate Research Fellowship under Grant DGE 2241144.}
    \thanks{All authors are with the Robotics Department, University of Michigan, Ann Arbor, MI, 48109, USA.
        {\tt\footnotesize\{wenzhet, jjomi, yixili, nfz, xiaonanh\}@umich.edu}}
}

\begin{document}

\maketitle
\thispagestyle{empty}
\pagestyle{empty}


\begin{abstract}
This paper presents a scalable, open-source visuotactile sensing system for tensegrity robots that enables six-axis wrench estimation and contact detection. The proposed endcap sensor integrates an elastomeric shell, a 3D-printed thermoplastic polyurethane (TPU) interface, and a rigid base housing an embedded camera and LED illumination ring. A novel gyroid-infill bonding technique is introduced to form a durable elastomer-TPU interface without adhesives, yielding a lightweight and modular design compatible with large-scale tensegrity structures. A tactile-to-wrench neural network maps shear vector fields to six-dimensional force and torque measurements. Experimental results demonstrate accurate and stable wrench estimation with a mean squared error (MSE) of 0.1531 on static validation data and out-of-domain generalization under dynamic motion. Furthermore, full-system integration on a 12 kg tensegrity robot confirms the sensor's ability to reliably identify ground contacts. The system substantially improves the practicality of tactile feedback for tensegrity robots, offering a low-cost, reproducible, and physically interpretable pathway toward contact-aware proprioception and state estimation.
Open source files are available at \href{https://github.com/Jonathan-Twz/tensegrity-gelfoot}{github.com/Jonathan-Twz/tensegrity-gelfoot}
\end{abstract}



\section{Introduction}
\input{subsections/1-introduction}

\section{Related Works}
\input{subsections/2-relatedWorks}

\section{Visuotactile Sensor Design}
\input{subsections/3-design}

\section{Materials and Methods}
\input{subsections/4-manuf}

\section{Contact Sensing}
\input{subsections/5-sw}

\section{Experimental Results}
\input{subsections/6-results}

\section{Conclusion and Future Works}
\input{subsections/7-conclusion}



\bibliographystyle{IEEEtran}
\bibliography{strings-abrv,ieee-abrv,ref}

\end{document}

%% file: subsections/1-introduction.tex
Tensegrity robots, characterized by a network of rigid struts and tensile cables, exhibit exceptional strength-to-weight ratios, resilience to impact, and the ability to navigate complex environments through shape morphing. 
Their flexible and collapsible structures make them promising candidates for applications in exploration, search-and-rescue, and soft robotics~\cite{skelton2009tensegrity, mi2024designvariablestiffnessquasidirect, bruceDesignEvolutionModular2014}.
However, achieving robust autonomous operation remains a challenge due to their high degrees of freedom and nonlinear dynamics, which complicate state estimation~\cite{shahTensegrityRobotics2022}.

Accurate state estimation is fundamental for feedback control and vision-based localization, providing the necessary priors for stable locomotion~\cite{campos2021orb, shan2020lio}.
Current proprioceptive methods for tensegrity robots typically rely on member-length measurements from resistive or capacitive-based strain sensors embedded in cables or structural elements~\cite{shahTensegrityRobotics2022, lu20226ndofposetrackingtensegrity, johnsonSensorTendonsSoft2022}. 
Some research features cable-spool encoder and inertial measurement units (IMUs)~\cite{tong2024tensegrityrobotproprioceptivestate, caluwaertsStateEstimationTensegrity2016b}. 
External sensing approaches, such as ranging sensors~\cite{caluwaertsStateEstimationTensegrity2016b} or external motion capture~\cite{lu20226ndofposetrackingtensegrity}, remain impractical for field deployment.

A major gap in current tensegrity state estimation literature is the lack of reliable direct ground contact measurement. Contact sensing can provide explicit constraints for future state estimators; knowing which endcap is anchored to the ground can transform a floating-base estimation problem into a more constrained formulation. While visuotactile sensing has proven effective in robotic manipulation~\cite{do_densetact_2023, taylor_gelslim30_2021,kuppuswamy_soft-bubble_2020,yuan_gelsight_2017}, its application to tensegrity robots remains largely unexplored. A scalable, integratable contact sensing system could therefore provide useful proprioceptive measurements without imposing excessive fabrication complexity or space constraints.

\begin{figure}[t]
    \centering
    \includegraphics[width=\linewidth]{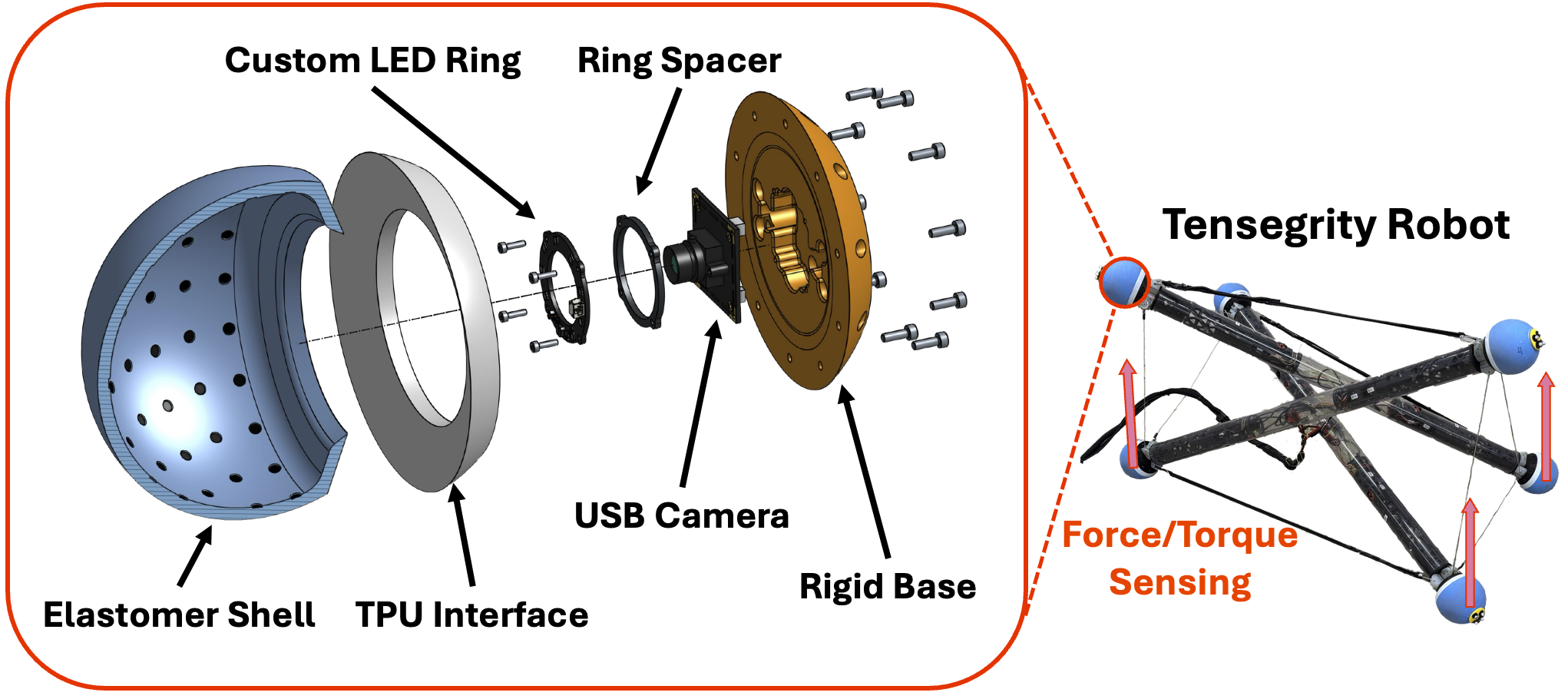}
    \caption{Exploded view of the visuotactile endcap sensor and its integration on a tensegrity robot. The sensor comprises an elastomer shell, TPU interface, custom LED ring, ring spacer, USB camera, and rigid base for distributed contact detection and contact wrench estimation.}
    \label{fig:sensor}
\end{figure}

The elastomeric visuotactile sensing system presented in this paper, shown in Fig. \ref{fig:sensor}, introduces key advancements for improving contact estimation in tensegrity robots. First, it leverages visuotactile sensing to enhance state estimation by directly sensing contact interactions, which is critical for feedback control and localization. Second, the sensor is designed with accessibility in mind, utilizing an open-source and low-cost fabrication process that enables straightforward replication and integration into various tensegrity platforms. Third, a novel method for bonding elastomer to 3D-printed structures is introduced, ensuring a durable interface. 
Finally, we evaluate a lightweight residual neural network that maps shear vector fields to contact forces and torques, enabling accurate, real-time contact wrench estimation for a tensegrity robot platform.
Together, these elements improve the practicality of deploying contact sensors on tensegrity robots and provide measurements that can support future contact-aware state estimation and control.


%% file: subsections/2-relatedWorks.tex
Within the field of tensegrity robots, contact sensing has received limited attention despite its potential to improve state estimation~\cite{tong2024tensegrityrobotproprioceptivestate}. Previous efforts by Booth et al. integrated molded resistive elements into elastomeric endcaps for boolean contact detection~\cite{boothSurfaceActuationSensing2021}, while Barkan et al. utilized force-sensing resistors (FSRs) to study human-robot interaction~\cite{BarkanHRITensegrity}. However, resistive sensors often suffer from drift and hysteresis, complicating long-term accuracy. Similarly, alternative modalities such as large-area capacitive sensors~\cite{PagolicapacitiveSensor} and Hall effect sensors~\cite{HarberMagneticTactileSensor, MohammadiMagneticSensor} face challenges related to fabrication complexity, signal drift~\cite{johnsonSensorTendonsSoft2022}, or magnetic interference. Furthermore, these methods typically struggle to provide the high-resolution shear and texture information offered by optical tactile sensing.

Our work advances tensegrity robot endcap-ground contact estimation by incorporating elastomeric visuotactile sensors that have recently been explored beyond robotic manipulation and into legged locomotion. These sensors typically consist of a deformable elastomer layer embedded with a visual marker pattern, which is observed by an integrated camera. Contact-induced deformations alter the marker pattern, enabling vision-based estimation of force, shear, and surface texture~\cite{do_densetact_2023}. Stone~\textit{et al.} introduced TacTip toes, a modular foot-mounted visuotactile sensor with an endoscopic camera module and a replaceable compliant tip~\cite{stone2020walking, chorley2009development, ward2018tactip}. Similarly, Song~\textit{et al.} introduced TacTID, with embedded elastic markers and demonstrated real-time terrain properties estimation during quadruped locomotion~\cite{song2024tactid}.
Depending on their optical and mechanical design, these sensors can further provide detailed tactile information, including object shape~\cite{taylor_gelslim30_2021}, texture~\cite{song2024tactid}, and force feedback~\cite{do_densetact_2023}.

Despite their advantages, visuotactile sensors remain challenging to scale and ruggedize for fielded, high-load robotic platforms. Their performance often depends on precise fabrication, including elastomer casting quality, marker consistency, and camera alignment. 
For example, TacTip requires special phosphorescent pigment painting to improve pin visibility, while TacTID adopts a dual-stage casting process to form embedded markers, increasing fabrication complexity and potential unit-to-unit variation. These requirements can hinder reproducible manufacturing at scale and reduce robustness under repeated impacts, abrasion, and environmental contamination.
Efforts like GelSlim 4.0 by Sipos~\textit{et al.} aim to improve reproducibility and manufacturability, but the fabrication process remains complex due to the number of unique components involved~\cite{sipos2024gelslim}. While these sensors demand greater computational resources than simpler alternatives, advancements in computing power and the increasing reliance on high-performance processors in autonomous systems have mitigated this constraint, making visuotactile sensing more viable for complex robotic applications.

%% file: subsections/3-design.tex
\begin{figure*}[ht]
    \centering
    \includegraphics[width=0.8\linewidth]{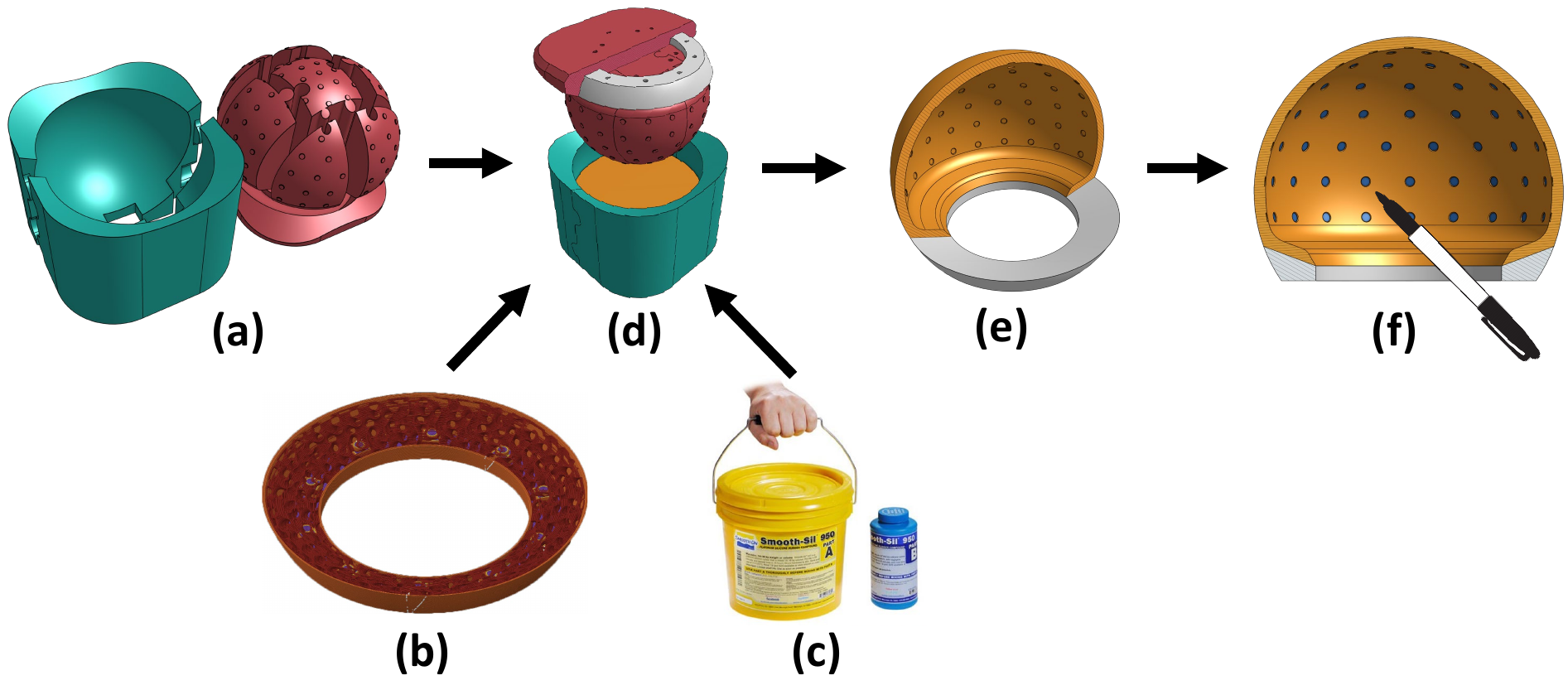}
    \caption{An overview of the manufacturing steps to produce one endcap. First, the (a) PLA mold and (b) TPU interface are 3D printed and assembled. Then, the (c) elastomer can be mixed and (d) poured into the mold. The (e) de-molded endcap can then be (f) dotted with a permanent marker.}
    \label{fig:manuf}
\end{figure*}

To meet the geometric and functional requirements of typical meter-scale tensegrity robots, the sensor, shown in Fig.~\ref{fig:sensor}, is designed with a 130 mm diameter to match the scale of the endcap. At this size, no commercial or non-commercial visuotactile sensors are readily available. The sensor consists of five core components: the elastomer shell, TPU interface, rigid base, LED light ring, and camera. All components are either easily fabricated or sourced off the shelf.

\subsection{Mechanical Design}

The visuotactile sensor consists of a compliant elastomer shell, a soft TPU interface, and a rigid base that houses the sensing (Fig.~\ref{fig:sensor}). The elastomer shell forms the primary contact surface with the environment, providing a soft yet durable interface capable of capturing fine deformations during contact. During casting, shallow dents are intentionally formed on the inner surface of the shell (Fig.~\ref{fig:manuf}a), and dots are painted onto these dents to serve as visual markers (Fig.~\ref{fig:manuf}f). This patterned surface enables the embedded camera to track deformation patterns for contact and shear force estimation. To ensure accurate transmission of contact-induced deformations, the shell is fabricated from a uniform elastomer material with consistent mechanical properties across its surface.

A soft TPU interface layer connects the elastomer shell to the rigid base, maintaining flexibility while providing mechanical robustness. The continuous-infill mechanical interlock described in Section IV securely joins the TPU and elastomer without compromising elasticity or requiring adhesives. The interface also contains embedded square nuts that establish a modular connection to the rigid base, allowing endcaps with different material properties—such as stiffness or shell thickness—to be easily interchanged.

The rigid base serves as the structural foundation of the sensor and houses both the miniature camera and the surrounding LED ring used for illumination (Fig.~\ref{fig:sensor}). It ensures the proper alignment of the optical components while protecting the internal hardware. Together, these components form a compact, robust, and reconfigurable visuotactile sensing module well suited for integration into tensegrity robots.

\subsection{Electrical Design}

The electrical system integrates a custom LED ring and a miniature USB camera to enable uniform illumination and high-resolution visual sensing within the visuotactile module (Fig:~\ref{fig:sensor}). The LED ring is implemented as a dedicated printed circuit board assembly (PCBA) that provides consistent lighting across the inner surface of the elastomer shell. It is powered through the camera’s 5 V output, which is directly connected to the USB 5 V rail. A current-controlled boost converter (TPS61165) raises the supply voltage to a nominal 25 V to drive the LED array (JB3030AWT) efficiently. The boost controller regulates the LED string current to 20 mA using a sense resistor placed at the end of the LED chain, maintaining stable brightness across operating conditions. With an overall efficiency of approximately 86\%, the LED ring draws less than 130 mA of current—well within the USB 2.0 power limit of 2.5 W—making it compact and self-contained without requiring an external power source.

The onboard camera (ELP-USBFHD01M, equipped with an OV2710 sensor) captures images of the patterned elastomer surface for contact and force estimation. It is fitted with a 180-degree fisheye lens to maximize the visible area of the shell and ensure uniform coverage of the tactile field. This wide field of view is particularly advantageous for tensegrity robots with non-spherical geometries, where ground contact frequently occurs along the side of the endcap rather than at its tip~\cite{tong2024tensegrityrobotproprioceptivestate}.

%% file: subsections/4-manuf.tex
The visuotactile endcap is fabricated from cast silicone, 3D-printed thermoplastic polyurethane (TPU), and a rigid base. The elastomer shell is made from Smooth-Sil™ 950 (Shore 50A), chosen for its high tear resistance and stiffness, which are necessary to support the large, meter-scale tensegrity robot. Softer materials can be substituted for smaller robots to improve tactile resolution. Although Smooth-Sil 950 has a high mixed viscosity ($\sim$35,000 cps), the mold design ensures complete filling without air entrapment. A lower-viscosity urethane (ReoFlex 60) was also evaluated but bonded irreversibly to the mold despite multiple release agents. The TPU interface is printed using Bambulabs TPU 95A filament, offering both elasticity and structural strength.

\begin{figure}[t]
    \centering
    \subfloat[]{
        \includegraphics[height=9.5em]{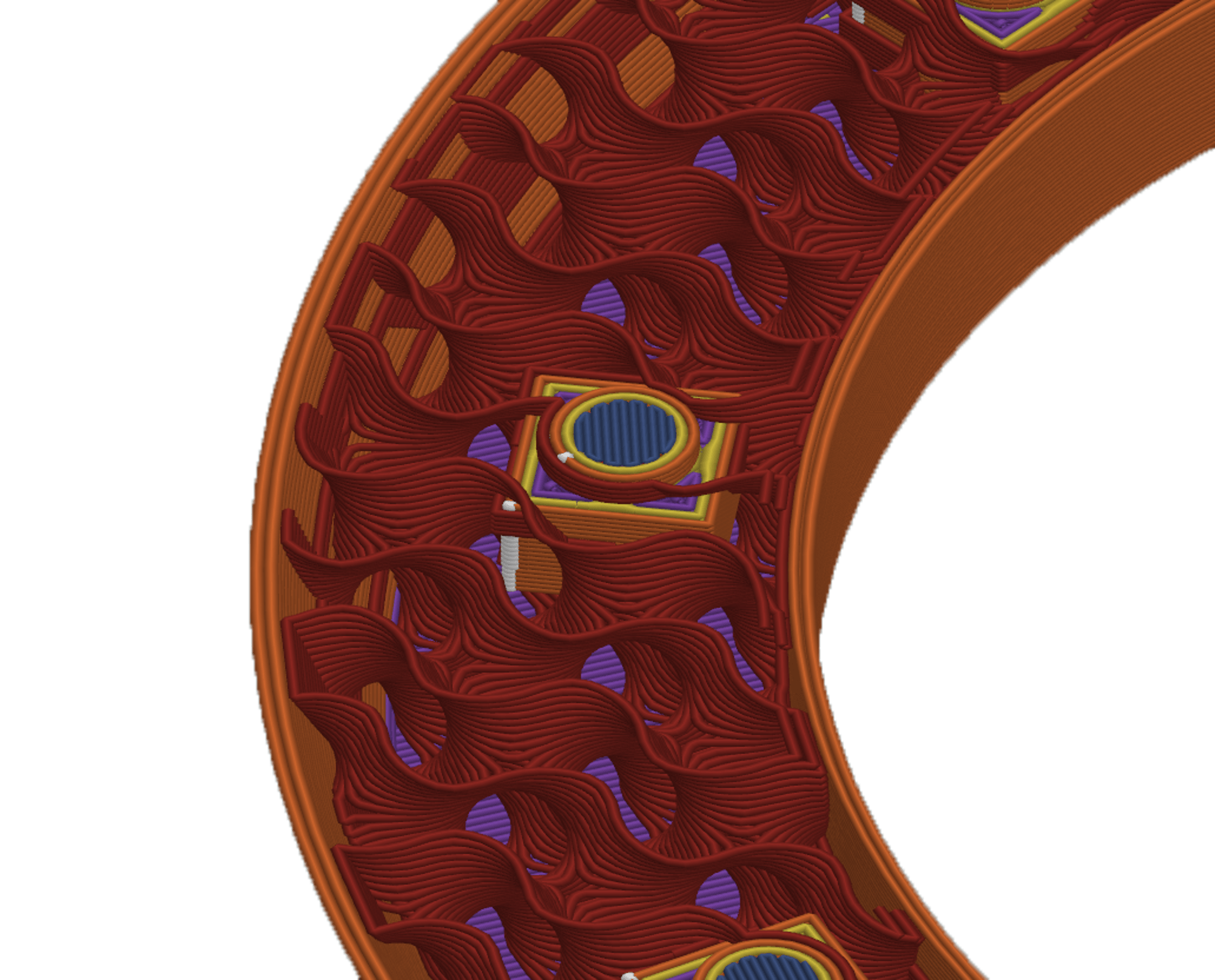}
        \label{fig:gyroid_slicer}
        }
    \subfloat[]{
        \includegraphics[height=9.5em]{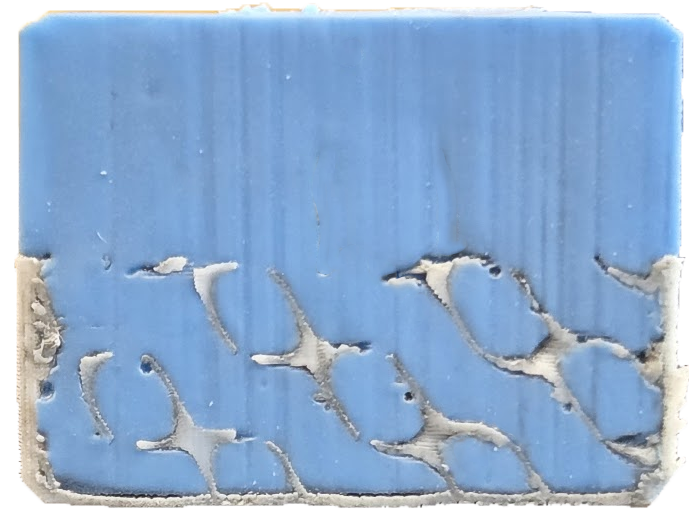}
        \label{fig:gyroid_irl}
        }
    \caption{Sectional view of gyroid infill from (a) 3D printing slicer software and (b) real life elastomer-TPU bonded sample. There is a 10\% infill in both of these samples.}
    \label{fig:gyroid}
\end{figure}

A continuous-infill mechanical bonding technique is used to join the elastomer shell and TPU interface. During molding, liquid silicone flows into the open gyroid infill of the 3D-printed TPU, creating a strong mechanical interlock once cured (Fig.~\ref{fig:gyroid}). This method eliminates the need for adhesives and produces a bond limited only by the materials' intrinsic strength. For the tensegrity endcap, a 10\% gyroid infill is used to balance bonding strength and material flexibility. 

The mold for casting the elastomer shell is fully 3D-printed and consists of two main parts with interlocking puzzle-like joints for easy demolding (Fig.~\ref{fig:manuf}a). The configurable CAD design allows the shell size and wall thickness to be adjusted through only two parameters. Dimples on the mold's interior create small dents on the shell surface (Fig.~\ref{fig:manuf}e), later filled with permanent marker ink to produce high-contrast visual features for deformation tracking (Fig.~\ref{fig:manuf}f). All mold components are printed in Polyerra PLA at low cost ($\sim$\$8.50) and are reusable for repeated fabrication cycles.

The manufacturing process begins by printing the mold and TPU interface with embedded square nuts for modular assembly. After applying a mold release agent (Ease Release 200, Smooth-On, Inc.), the elastomer mixture is prepared, degassed, and poured into the lower mold half before the upper section—holding the TPU interface—is inserted (Fig:~\ref{fig:manuf}b-d). The assembly is clamped and cured either overnight at room temperature or in an oven for faster processing. Once cured, the mold separates cleanly thanks to its segmented design, yielding a finished endcap ready for marker application.

The parametric, modular design enables scalable fabrication across multiple robot sizes without changing the sensing hardware. The same camera and illumination setup can be used to achieve consistent visuotactile performance across different tensegrity robot platforms.

%% file: subsections/5-sw.tex
\begin{figure}[htbp]
    \centering
    \includegraphics[width=\linewidth]{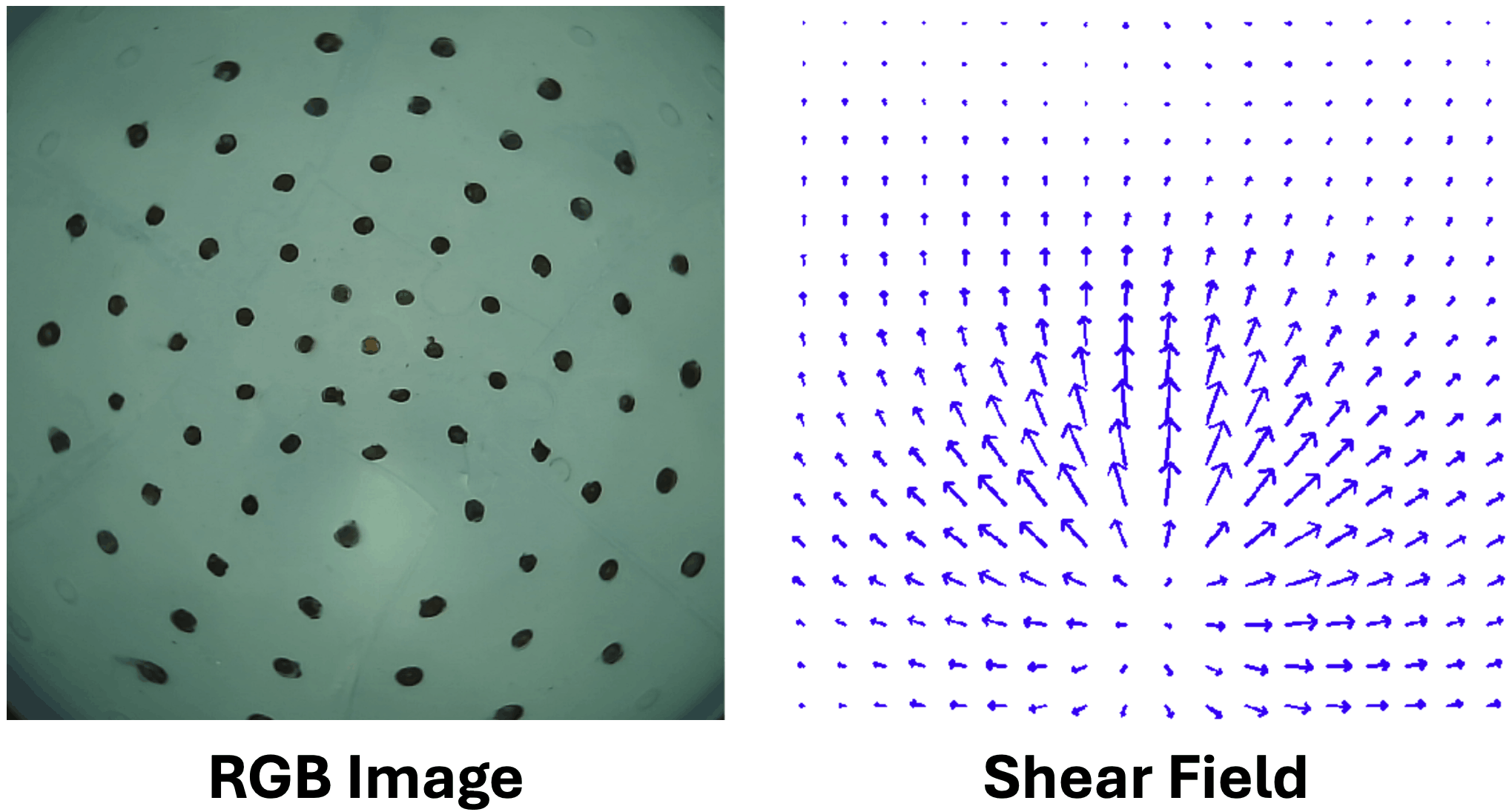}
    \caption{Endcap camera image and the corresponding $18 \times 18$ shear field generated by shear field estimation methods from Gelslim4.0 \cite{sipos2024gelslim}, during contact near the lower edge of the frame.}
    \label{fig:sensorData}
\end{figure}

We train a tactile-to-wrench neural network that maps a dense gel-foot vector field to the 6D contact wrench, and then derive a binary contact indicator from the predicted wrench. The input at each time step is a two-channel shear vector field\, $\mathbf{V} \in \mathbb{R}^{2\times H\times W}$ (as shown in Fig.~\ref{fig:sensorData}) computed using shear-field estimation methods from GelSlim 4.0 \cite{sipos2024gelslim} based on optical flow, and the target is the wrench $\mathbf{w} = [F_x,F_y,F_z,\,T_x,T_y,T_z]^\top \in \mathbb{R}^6$ measured by a force/torque sensor.

\subsection{Dataset and preprocessing} 
    \label{sec: dataset-and-preprocessing}
    We collect a contact-wrench dataset using a KUKA Med R820 industrial robot with a supporting plane mounted on an ATI Gamma 6-axis force/torque sensor fixed on a table, by executing scripted static down-presses with variation in orientation, force, and motion direction. RGB frames at 30 Hz with a nominal resolution of $640\times480$, center-cropped to a square field of view, and downsampled to $200\times200$ before optical-flow processing. We then compute a two-channel shear field of shape $(2,30,30)$ in real time using the same methods described in Gelslim 4.0 \cite{sipos2024gelslim}. We also transform force and torque from the ATI Gamma F/T sensor frame to the endcap sensor center frame. Shear-field and wrench messages are recorded as synchronized pairs using their ROS timestamps with an approximate synchronization tolerance of 50 ms. The final dataset contains 34,368 labeled samples.
    
\begin{figure}[htbp]
        \centering
        \includegraphics[width=0.95\linewidth]{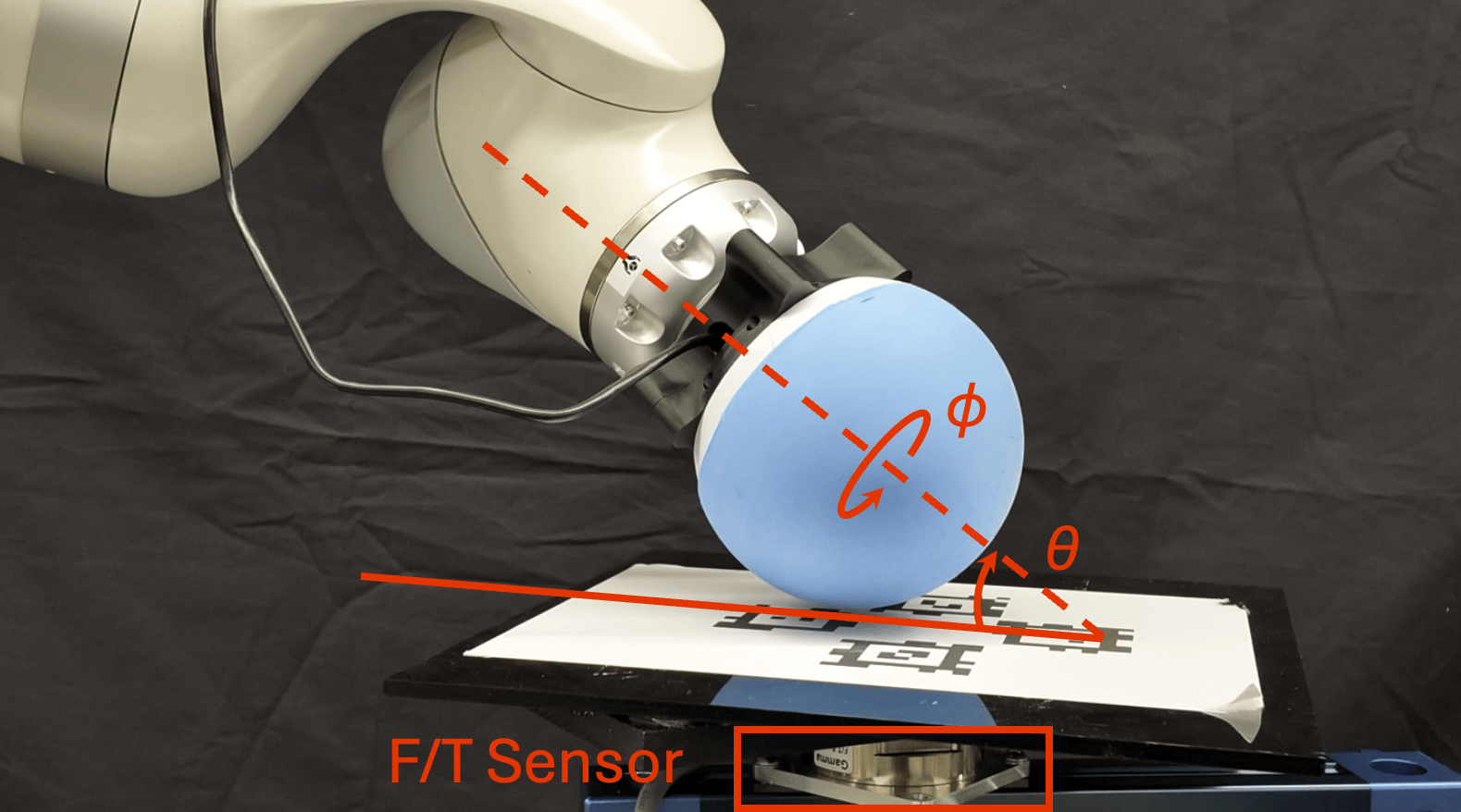}
        \caption{Endcap visuotactile sensor characterization with KUKA Med R820 industrial robot arm and ATI Gamma 6-axis force/torque sensor. The sensor approach orientation is parameterized by tilt angle $\theta$ and twist angle $\phi$ about the axial direction. We sample $\theta$ from $10^{\circ}$ to $50^{\circ}$ in $10^{\circ}$ steps and $\phi$ every $45^{\circ}$ from $0^{\circ}$ to $360^{\circ}$.}
        \label{fig:sensor-contact-frame}
     \end{figure}

    The experiment setup is shown in Fig.~\ref{fig:sensor-contact-frame}. For each trial, the end-effector poses the tool at a prescribed approach orientation parameterized by tilt angle $\theta$ and twist angle $\phi$ :
    \begin{equation}
        \theta \in \{10^{\circ},20^{\circ},30^{\circ},40^{\circ},50^{\circ}\},\\
        \phi \in \{0^{\circ},45^{\circ},90^{\circ},\ldots,315^{\circ}\}.
    \end{equation}
    At each $(\theta,\phi)$ contact orientation, we first apply a sequence of normal-force setpoints spanning $[5,35]$ N while maintaining the commanded orientation, then draw a circle on the supporting plane to cover more frictional cases. 
    
\begin{figure}[htbp]
        \centering
        \includegraphics[width=\linewidth]{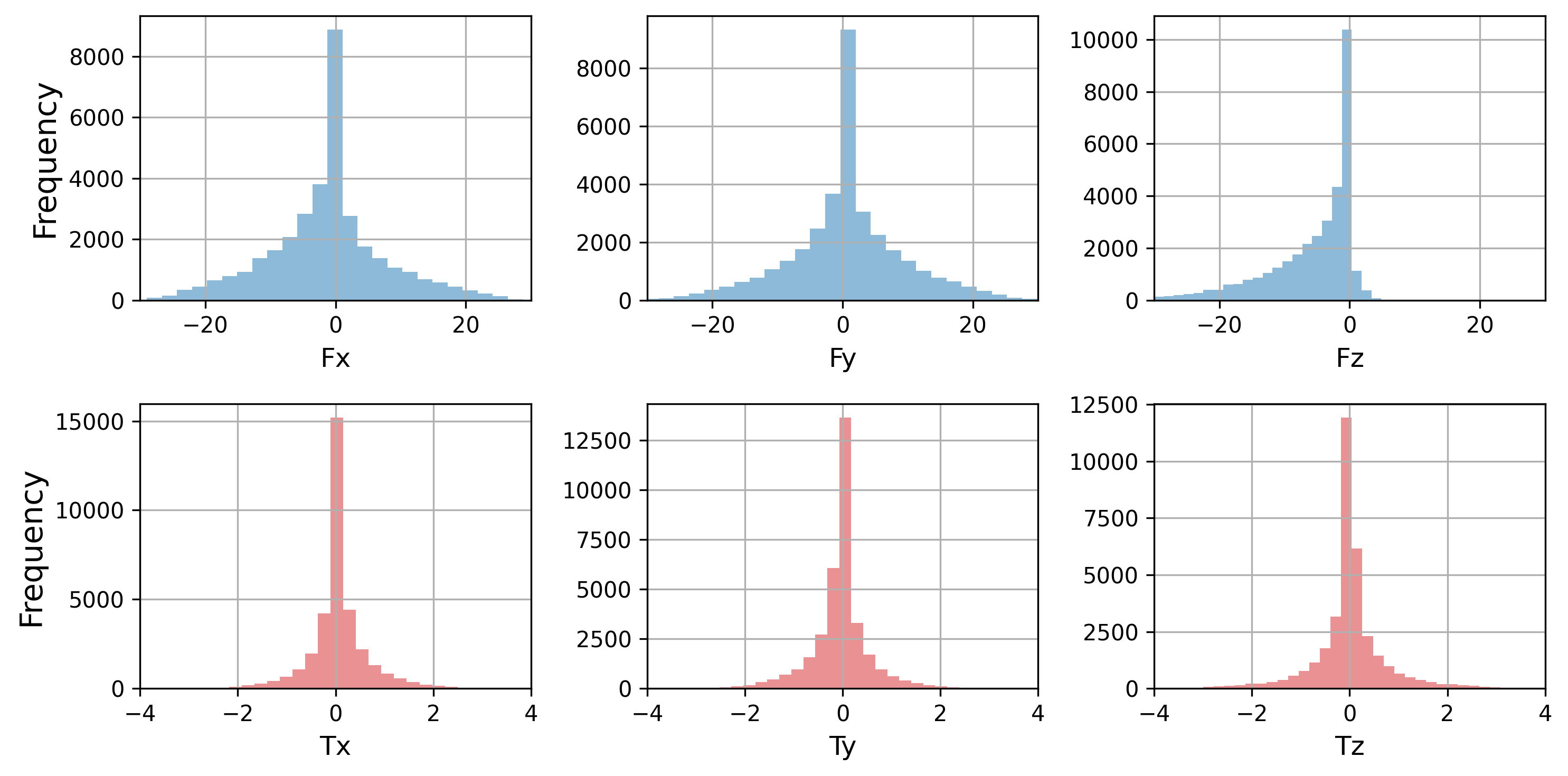}
        \caption{Histogram of training set wrench label over six axes.}
        \label{fig:training-data-distribuation}
    \end{figure}

    We visualize the collected training dataset as a histogram, as shown in Fig.~\ref{fig:training-data-distribuation}. The distribution is intentionally dominated by low-to-moderate contact forces because each trial includes approach, light contact, and unloading phases in addition to high-force presses.
    
\subsection{Model}
    We adopt a deep residual Multi-Layer Perceptron (MLP)~\cite{he2016deep, xie2017aggregated} as the shear-to-wrench mapping function. The architecture consists of an initial projection layer followed by two residual blocks with a hidden dimension of 64, and a final linear regression head. The full forward pass is defined as:
    \begin{align}
        \hat{\mathbf{w}} &= f_\theta\big(\mathrm{vec}(\mathbf{V})\big), \\
        f_\theta &=\underbrace{\mathrm{FC}\!\to\!\mathrm{ReLU}\!\to\!\mathrm{FC}}_{\text{ResBlock}}\times K_\mathrm{blk}\; \to\; \mathrm{FC}_{6},
    \end{align}
    where $\mathrm{vec}(\mathbf{V})$ denotes the flattened input shear field vector and $K_{blk}$ is the number of residual blocks. The residual architecture facilitates gradient flow during training, allowing the efficient learning of the nonlinear mapping between shear field and wrench outputs.
    
\subsection{Training}
    We split the dataset into 70\% as training set, 15\% as test set, and 15\% as validation set. Here, we use MSE loss,
    \begin{equation}
    	\mathcal{L}(\theta)=\tfrac{1}{N}\sum_{n=1}^{N} \lVert\hat{\mathbf{w}}^{(n)}-\mathbf{w}^{(n)}\rVert_2^2,
    \end{equation}
    which minimizes the mean squared error between the predicted wrench $\hat{\mathbf{w}}$ and the ground-truth $\mathbf{w}$ across the batch of $N$ samples.
    The network is trained with Adam optimizer at a learning rate of $10^{-3}$ and a batch size of 32 for 100 epochs on an NVIDIA RTX~4070Ti GPU. The total training time is approximately 120s. We monitor both training and validation losses at each epoch and retain the checkpoint with the lowest validation error.

\subsection{Inference}
    From the predicted wrench, we derive detection signals needed for control and estimation:
        \begin{align}
        	\hat{\mathbf{F}} &=[\hat{F}_x,\hat{F}_y,\hat{F}_z]^\top, \\
            c &= \mathbb{I}\big[\lVert\hat{\mathbf{F}}\rVert_2 > \tau_F\big],
        \end{align}
    where $c\in\{0,1\}$ indicates contact. We select threshold $\tau_F$ from a no-contact calibration interval as the smallest value that suppresses false positive contacts in the suspended sensor state; for the six-endcap robot experiment, we use $\tau_F=0.2$ N for all endcaps. This wrench-based approach yields contact cues while remaining agnostic to the exact gel geometry.


%% file: subsections/6-results.tex



\subsection{TPU and Elastomer Interface Testing}
To ensure reliable integration of TPU components with elastomeric endcaps, it is critical to evaluate the strength and consistency of different bonding methods. 
In the context of tensegrity robots, these bonds must withstand both tensile and shear forces encountered during locomotion and environmental interactions. 
Tensile strength is a key factor in preventing delamination under normal and impact loads, while shear strength determines resistance to peeling and sliding forces. 
To assess these factors, we conducted tensile and peel tests using a universal testing machine (Instron 68SC-5). 
Test specimens were fabricated using the same Smooth-Sil 950 elastomer and TPU 95A materials as the endcap to ensure results accurately reflect real-world performance.
Four different bonding methods were tested: mechanical bonding, Sil-Poxy\texttrademark, Ure-Bond\texttrademark~II, and 3M\texttrademark 80 spray adhesive. 
Sil-Poxy\texttrademark and Ure-Bond\texttrademark~II are commonly used in soft robotics applications for soft material bonding, and 3M\texttrademark~80 spray adhesive is specifically designed for rubber adhesion.
Of these four elastomeric adhesives, Ure-Bond\texttrademark ~II and 3M\texttrademark~80 did not bond with the TPU and, as a result, could not be experimentally tested.

\subsubsection{Tensile}
\begin{figure}[t]
    \centering
    \subfloat[]{
        \includegraphics[width=0.49\linewidth]{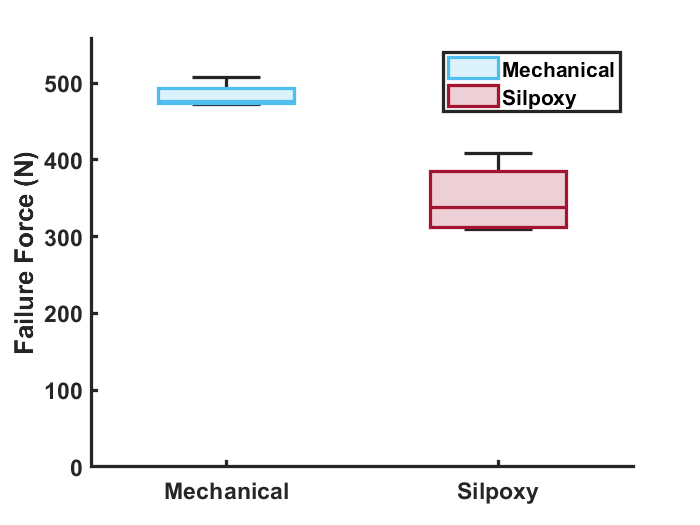}
        \label{fig:tensileBox}
        }
    \subfloat[]{
        \includegraphics[width=0.49\linewidth]{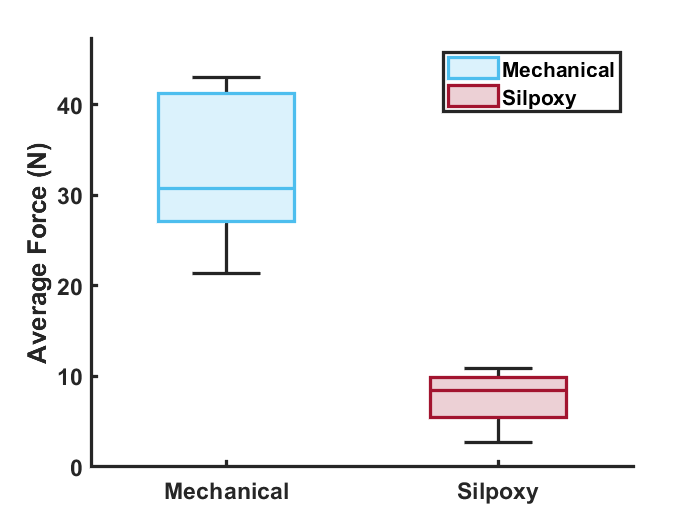}
        \label{fig:peelBox}
        }
    \caption{Box plot of (a) max tensile and (b) average peel forces endured by mechanical and Sil-Poxy bonding of Smooth-Sil 950 and TPU/.}
    \label{fig:mechTest}
\end{figure}

To evaluate the tensile strength, a pull-to-failure test was used to compare mechanical bonding and Sil-Poxy adhesive. 
The test specimens consisted of 25 mm $\times$ 25 mm $\times$ 50 mm rectangular prisms of elastomer bonded to TPU blocks of the same dimensions. 
For mechanical bonding, 25 mm of interference was utilized, accurately representing a cross-section of the robot's endcap interface. 
Each specimen was subjected to tensile loading at a rate of 10 mm/min until failure. 
The results, presented in Fig.~\ref{fig:tensileBox}, indicate that mechanical bonding provides approximately 25\% higher tensile strength than Sil-Poxy while exhibiting significantly lower variation.

\subsubsection{Shear}
Shear strength was assessed using a peel test based on ASTM D1876. 
Test specimens were 25 mm $\times$ 300 mm $\times$ 25 mm, with an unbonded length of 100 mm—slightly increased from the standard 76 mm to accommodate the thicker specimens. 
The Instron machine pulled at a rate of 100 mm/min until complete separation occurred. 
The results show that mechanical bonding is approximately 300\% stronger than Sil-Poxy; however, greater variation was observed due to material failure occurring before the mechanical bond could fail in some trials.

\subsection{Static Contact Wrench Validation}
    
    We evaluate the tactile-to-wrench model on the held-out validation set described in Section~\ref{sec: dataset-and-preprocessing}. The mean squared error (MSE) for wrench $\mathbf{w}$ is 0.1314 on the training set, 0.1513 on the test set, and 0.1531 on the validation set. The validation error remains close to the training loss, suggesting that the model is not strongly overfitting the sampled static contact distribution. These aggregate MSE values are used only as a compact summary; because the wrench combines forces in N and torques in N$\cdot$m, per-axis force and torque errors should be interpreted separately when assessing physical accuracy.

\begin{figure}[htbp]
        \centering
        \includegraphics[width=\linewidth]{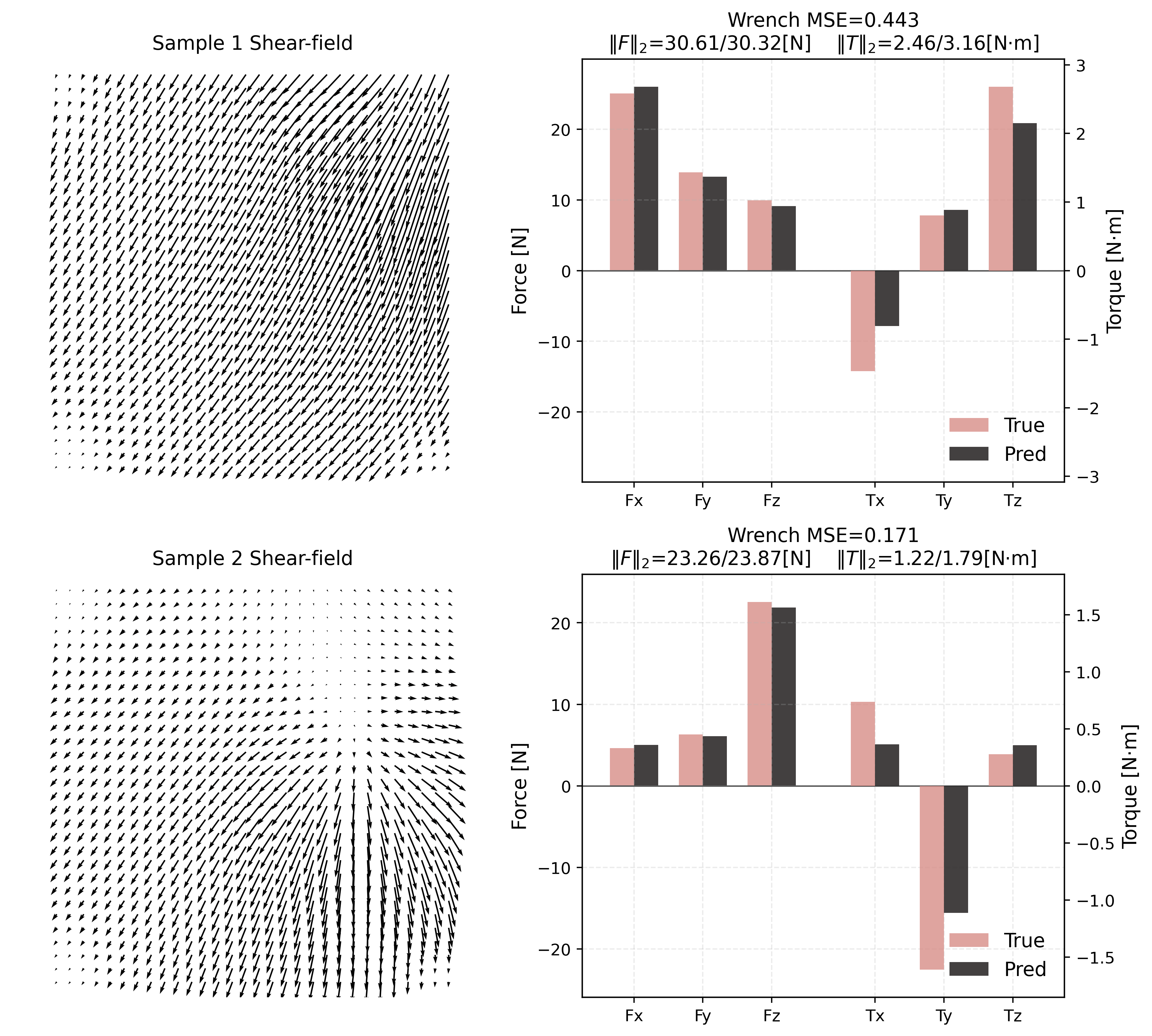}
        \caption{Static contact wrench validation on held-out $(\theta,\phi)$ poses. Rows show example shear fields and corresponding predicted vs. ground-truth wrenches.}
        \label{fig:static-validation}
    \end{figure}

    Figure~\ref{fig:static-validation} presents two representative validation examples: Sample 1 exhibits a large tangential force, while Sample 2 features dominant normal force. 
    The left column visualizes the measured shear fields, and the right column compares the predicted wrench components against the ground-truth labels. The model reproduces the dominant force direction and magnitude in these representative samples, with close alignment in the normal force $F_z$ and reasonable agreement in lateral force components ($F_x, F_y$). Torque prediction is more sensitive to small spatial errors in the shear field, particularly near the gel boundary where optical distortions and illumination nonuniformity introduce local biases. 
    Overall, the results demonstrate that the residual MLP can infer physically meaningful contact wrench patterns from the observed shear field, with force estimation being more robust than torque estimation.
    
\subsection{Dynamic Contact Wrench Validation}
    To assess behavior beyond the quasi-static training distribution, we execute circular trajectories with the robot arm and continuously predict the wrench online. 
    The resulting mean squared error (MSE) is 2.67, which is significantly higher than the static contact prediction error of 0.1531. This increase indicates that the current model should not be interpreted as fully dynamic-calibrated; rather, the test probes how well a static shear-to-wrench model transfers to contact with motion, sliding, and latency.
    
\begin{figure}[htbp]
        \centering
        \includegraphics[width=\linewidth]{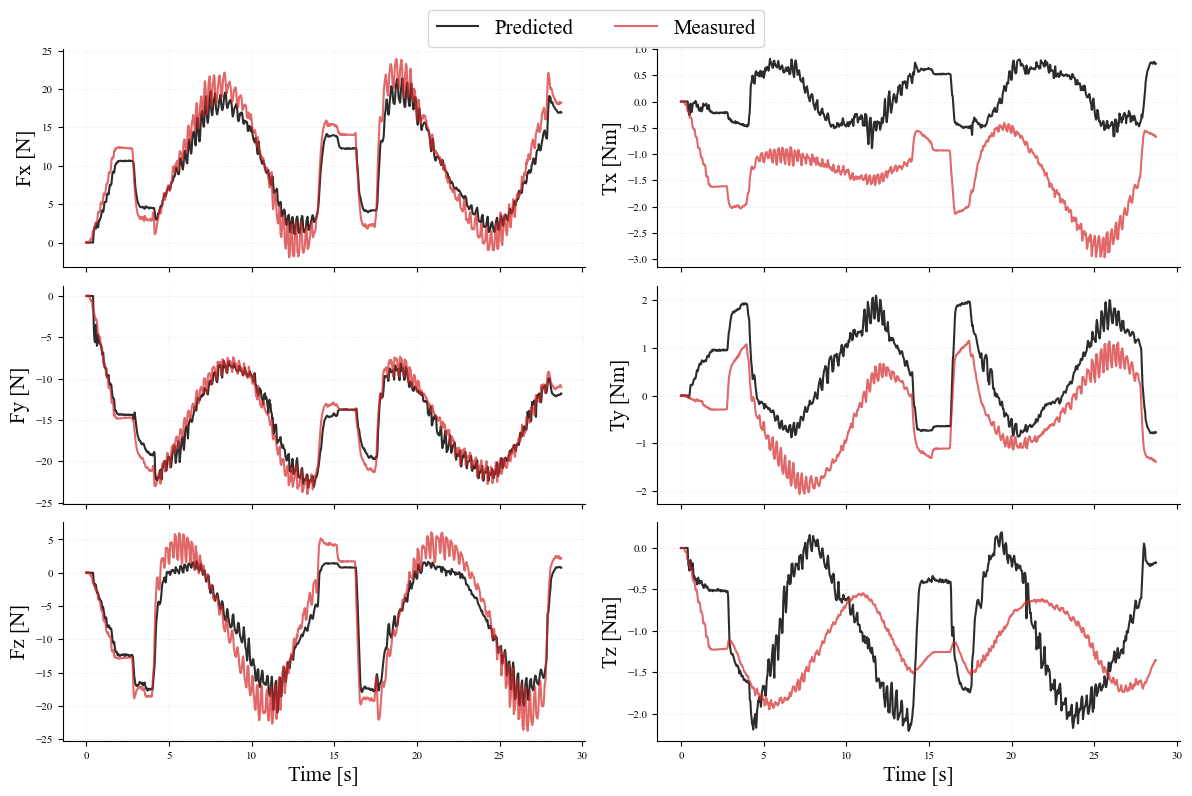}
        \caption{Temporal comparison of predicted and measured 6D contact wrenches during dynamic(out-of-domain) contact.}
        \label{fig:dynamic-validation}
    \end{figure}

    We also visualize the predicted and ground-truth wrenches over time in Fig.~\ref{fig:dynamic-validation} and Supplementary Video. The model tracks the main temporal trends of the force components, but the predictions are less accurate than in the static validation set. Torque estimation exhibits higher variability and larger phase lag, attributable to the finite camera frame rate, processing latency in shear-field estimation, and the lack of temporal history in the current residual MLP. Slight underestimation of peak $\lVert\mathbf{F}_t\rVert$ is observed, consistent with partial saturation of shear cues during sliding contact. 

    Despite these limitations, the model preserves coherent force trends that remain useful for contact detection. Notably, force estimation remains more robust than torque prediction under out-of-distribution dynamic motion. Future versions should incorporate higher-frame-rate imaging, timestamp compensation, and temporal models to reduce phase lag and improve torque prediction during sliding contacts.

\subsection{Tensegrity Robot Testing}

To demonstrate the practical applicability of the proposed sensor, we integrated six visuotactile endcaps onto the 12.0~kg, three-bar tensegrity robot, and evaluated its ability to estimate ground reaction force trends and detect contact events in real time. 

The robot was placed in four steady configurations: suspended with no contact, single contact, two-point contact, and stable three-point support. We recorded all six endcap force estimates simultaneously and assigned contact labels manually from video. 

\begin{figure*}[htbp]
    \centering
    \includegraphics[width=0.9\linewidth]{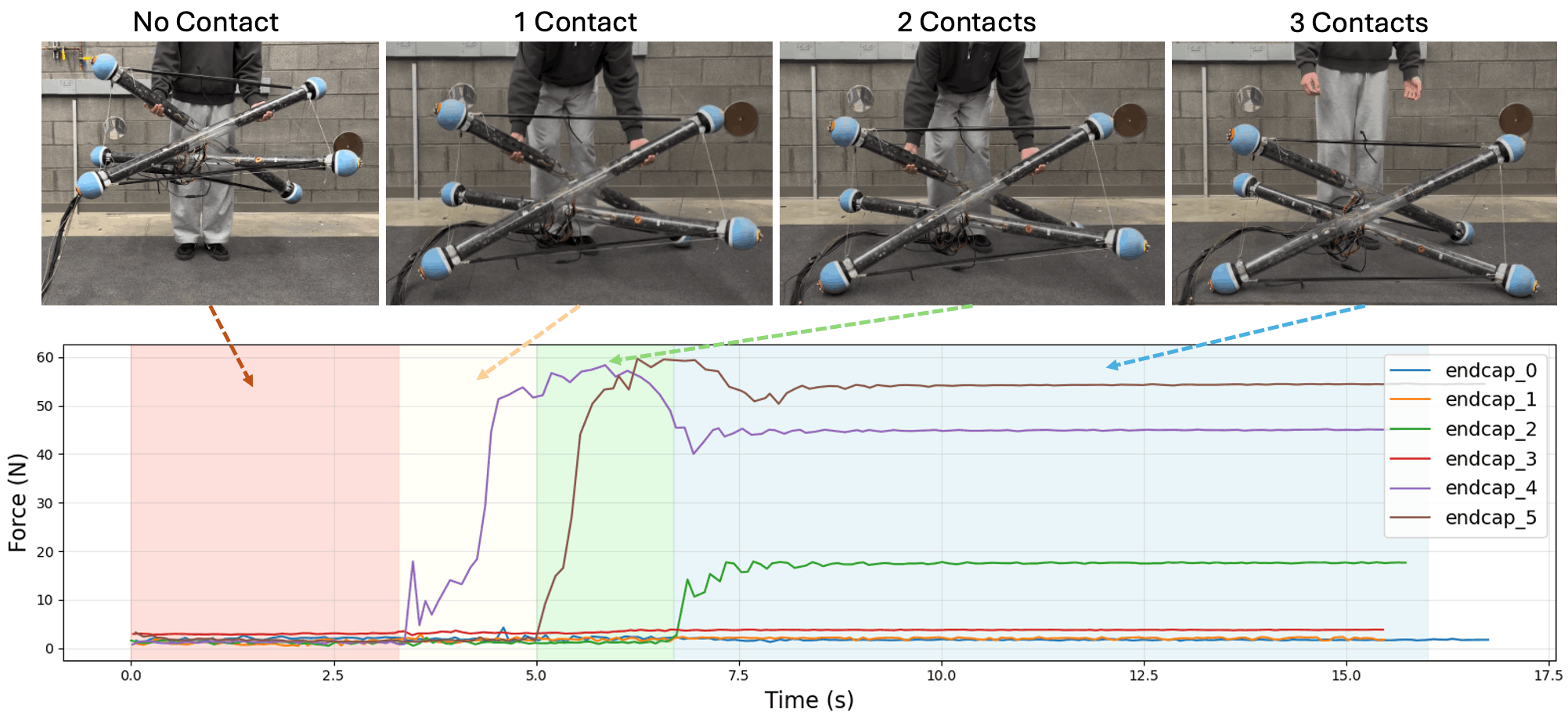}
    \caption{Validation of endcap visuotactile sensing on a tensegrity robot under varying ground-contact conditions. The robot is sequentially placed into configurations with 0, 1, 2, and 3 simultaneous ground contacts (top). The corresponding time series (bottom) shows the estimated contact force for each endcap: forces rise sharply for endcaps in contact and remain near zero for endcaps without contact.}
    \label{fig:tensegrity-experiments}
\end{figure*}

As shown in Fig.~\ref{fig:tensegrity-experiments}, contacting endcaps show sharp force increases while non-contacting endcaps remain near zero. The predicted active endcap set matches the manual labels in all four configurations. These results show that the sensor array provides reliable binary contact cues for future contact-aware state estimation, while full robot-level wrench validation remains future work.

%% file: subsections/7-conclusion.tex
This paper presents a scalable visuotactile endcap sensor for tensegrity robots that combines mechanical robustness, sensing precision, and ease of fabrication. The design integrates a compliant Smooth-Sil™ 950 elastomer shell, a TPU interface, and a rigid base housing a fisheye camera with a custom LED ring for uniform illumination. A novel continuous-infill bonding method enables direct interlocking between liquid silicone and 3D-printed TPU, eliminating adhesives and producing a stronger, more consistent interface. The modular architecture supports rapid reconfiguration and potential scalability across different robot sizes.

The tactile-to-wrench residual neural network maps shear deformation to 6D contact wrench estimates, achieving low prediction error in static contacts and useful force estimation during dynamic contacts. Torque estimation exhibits greater variability due to camera latency, limited shear-field resolution, and the absence of temporal modeling, so torque predictions should be interpreted more cautiously than force predictions. Furthermore, successful integration on a 12 kg tensegrity robot confirmed the sensor's ability to detect ground contact events in real-time, validating its potential for future contact-aware state estimation.

Future work will focus on improving wrench prediction accuracy through higher-frame-rate imaging, timestamp compensation, temporal networks, and integration with full-body contact-aware state estimation on tensegrity robots. Additionally, we plan to extend validation to unstructured terrains, diverse contact geometries, repeated impact conditions, and longer-term durability tests, moving beyond flat-surface experiments to better reflect real-world deployment scenarios.
The open-source design aims to promote reproducibility and broader adoption of visuotactile sensing in compliant robotic systems.

%% file: ref.bib
@inproceedings{bruceDesignEvolutionModular2014,
  title = {Design and Evolution of a Modular Tensegrity Robot Platform},
  booktitle = {2014 {{IEEE International Conference}} on {{Robotics}} and {{Automation}} ({{ICRA}})},
  author = {Bruce, Jonathan and Caluwaerts, Ken and Iscen, Atil and Sabelhaus, Andrew P. and SunSpiral, Vytas},
  year = {2014},
  month = may,
  pages = {3483--3489},
  publisher = {IEEE},
  address = {Hong Kong, China},
  doi = {10.1109/ICRA.2014.6907361},
  urldate = {2023-09-08},
  isbn = {978-1-4799-3685-4}
}

@inproceedings{johnsonSensorTendonsSoft2022,
  title = {Sensor {{Tendons}} for {{Soft Robot Shape Estimation}}},
  booktitle = {2022 {{IEEE Sensors}}},
  author = {Johnson, William R. and Agrawala, Anjali and Huang, Xiaonan and Booth, Joran and {Kramer-Bottiglio}, Rebecca},
  year = {2022},
  month = oct,
  pages = {1--4},
  publisher = {IEEE},
  address = {Dallas, TX, USA},
  doi = {10.1109/SENSORS52175.2022.9967136},
  urldate = {2023-09-03},
  isbn = {978-1-66548-464-0}
}

@article{shahTensegrityRobotics2022,
  title = {Tensegrity {{Robotics}}},
  author = {Shah, Dylan S. and Booth, Joran W. and Baines, Robert L. and Wang, Kun and Vespignani, Massimo and Bekris, Kostas and {Kramer-Bottiglio}, Rebecca},
  year = {2022},
  month = aug,
  journal = {Soft Robotics},
  volume = {9},
  number = {4},
  pages = {639--656},
  issn = {2169-5172, 2169-5180},
  doi = {10.1089/soro.2020.0170},
  urldate = {2024-07-23},
  copyright = {https://www.liebertpub.com/nv/resources-tools/text-and-data-mining-policy/121/},
  langid = {english}
}

@inproceedings{caluwaertsStateEstimationTensegrity2016b,
  title = {State Estimation for Tensegrity Robots},
  booktitle = {2016 {{IEEE International Conference}} on {{Robotics}} and {{Automation}} ({{ICRA}})},
  author = {Caluwaerts, Ken and Bruce, Jonathan and Friesen, Jeffrey M. and SunSpiral, Vytas},
  year = {2016},
  month = may,
  pages = {1860--1865},
  publisher = {IEEE},
  address = {Stockholm, Sweden},
  doi = {10.1109/ICRA.2016.7487331},
  urldate = {2024-07-25},
  isbn = {978-1-4673-8026-3}
}

@article{boothSurfaceActuationSensing2021,
  title = {Surface {{Actuation}} and {{Sensing}} of a {{Tensegrity Structure Using Robotic Skins}}},
  author = {Booth, Joran W. and {Cyr-Choini{\`e}re}, Olivier and Case, Jennifer C. and Shah, Dylan and Yuen, Michelle C. and {Kramer-Bottiglio}, Rebecca},
  year = {2021},
  month = oct,
  journal = {Soft Robotics},
  volume = {8},
  number = {5},
  pages = {531--541},
  issn = {2169-5172, 2169-5180},
  doi = {10.1089/soro.2019.0142},
  urldate = {2024-07-25},
  copyright = {https://www.liebertpub.com/nv/resources-tools/text-and-data-mining-policy/121/},
  langid = {english}
}

@InProceedings{lu20226ndofposetrackingtensegrity,
    author={Shiyang Lu and William R. Johnson III au2 and Kun Wang and Xiaonan Huang and Joran Booth and Rebecca Kramer-Bottiglio and Kostas Bekris},
    editor="Billard, Aude
    and Asfour, Tamim
    and Khatib, Oussama",
    title="6N-DoF Pose Tracking for Tensegrity Robots",
    booktitle="Robotics Research",
    year="2023",
    publisher="Springer Nature Switzerland",
    address="Cham",
    pages="136--152",
    isbn="978-3-031-25555-7"
}

@book{skelton2009tensegrity,
  title={Tensegrity systems},
  author={Skelton, Robert E and De Oliveira, Mauricio C},
  volume={1},
  year={2009},
  publisher={Springer}
}

@misc{tong2024tensegrityrobotproprioceptivestate,
      title={Tensegrity Robot Proprioceptive State Estimation with Geometric Constraints}, 
      author={Wenzhe Tong and Tzu-Yuan Lin and Jonathan Mi and Yicheng Jiang and Maani Ghaffari and Xiaonan Huang},
      year={2024},
      eprint={2410.24226},
      archivePrefix={arXiv},
      primaryClass={cs.RO},
      url={https://arxiv.org/abs/2410.24226}, 
}

@article{campos2021orb,
  title={Orb-slam3: An accurate open-source library for visual, visual--inertial, and multimap slam},
  author={Campos, Carlos and Elvira, Richard and Rodr{\'\i}guez, Juan J G{\'o}mez and Montiel, Jos{\'e} MM and Tard{\'o}s, Juan D},
  journal={IEEE transactions on robotics},
  volume={37},
  number={6},
  pages={1874--1890},
  year={2021},
  publisher={IEEE}
}

@inproceedings{shan2020lio,
  title={{LIO-SAM}: Tightly-coupled lidar inertial odometry via smoothing and mapping},
  author={Shan, Tixiao and Englot, Brendan and Meyers, Drew and Wang, Wei and Ratti, Carlo and Rus, Daniela},
  booktitle=C-IROS,
  pages={5135--5142},
  year={2020},
  organization={IEEE}
}

@inproceedings{do_densetact_2023,
	address = {London, United Kingdom},
	title = {{DenseTact} 2.0: {Optical} {Tactile} {Sensor} for {Shape} and {Force} {Reconstruction}},
	copyright = {https://doi.org/10.15223/policy-029},
	isbn = {979-8-3503-2365-8},
	shorttitle = {{DenseTact} 2.0},
	url = {https://ieeexplore.ieee.org/document/10161150/},
	doi = {10.1109/ICRA48891.2023.10161150},
	urldate = {2024-09-18},
	booktitle = {2023 {IEEE} {International} {Conference} on {Robotics} and {Automation} ({ICRA})},
	publisher = {IEEE},
	author = {Do, Won Kyung and Jurewicz, Bianca and Kennedy, Monroe},
	month = may,
	year = {2023},
	pages = {12549--12555},
}

@misc{taylor_gelslim30_2021,
	title = {{GelSlim3}.0: {High}-{Resolution} {Measurement} of {Shape}, {Force} and {Slip} in a {Compact} {Tactile}-{Sensing} {Finger}},
	copyright = {Creative Commons Attribution Non Commercial Share Alike 4.0 International},
	shorttitle = {{GelSlim3}.0},
	url = {https://arxiv.org/abs/2103.12269},
	doi = {10.48550/ARXIV.2103.12269},
	urldate = {2024-09-18},
	publisher = {arXiv},
	author = {Taylor, Ian and Dong, Siyuan and Rodriguez, Alberto},
	year = {2021},
	note = {Version Number: 1},
}

@article{yuan_gelsight_2017,
	title = {{GelSight}: {High}-{Resolution} {Robot} {Tactile} {Sensors} for {Estimating} {Geometry} and {Force}},
	volume = {17},
	copyright = {https://creativecommons.org/licenses/by/4.0/},
	issn = {1424-8220},
	shorttitle = {{GelSight}},
	url = {https://www.mdpi.com/1424-8220/17/12/2762},
	doi = {10.3390/s17122762},
	language = {en},
	number = {12},
	urldate = {2025-02-14},
	journal = {Sensors},
	author = {Yuan, Wenzhen and Dong, Siyuan and Adelson, Edward},
	month = nov,
	year = {2017},
	pages = {2762},
}

@inproceedings{kuppuswamy_soft-bubble_2020,
	address = {Las Vegas, NV, USA},
	title = {Soft-bubble grippers for robust and perceptive manipulation},
	copyright = {https://ieeexplore.ieee.org/Xplorehelp/downloads/license-information/IEEE.html},
	isbn = {978-1-7281-6212-6},
	url = {https://ieeexplore.ieee.org/document/9341534/},
	doi = {10.1109/IROS45743.2020.9341534},
	urldate = {2025-02-14},
	booktitle = {2020 {IEEE}/{RSJ} {International} {Conference} on {Intelligent} {Robots} and {Systems} ({IROS})},
	publisher = {IEEE},
	author = {Kuppuswamy, Naveen and Alspach, Alex and Uttamchandani, Avinash and Creasey, Sam and Ikeda, Takuya and Tedrake, Russ},
	month = oct,
	year = {2020},
	pages = {9917--9924},
}

@INPROCEEDINGS{BarkanHRITensegrity,
  author={Barkan, Andrew R. and Padmanabha, Akhil and Tiemann, Sala R. and Lee, Albert and Kanter, Matthew P. and Agarwal, Yash S. and Agogino, Alice M.},
  booktitle={2021 IEEE International Conference on Robotics and Automation (ICRA)}, 
  title={Force-Sensing Tensegrity for Investigating Physical Human-Robot Interaction in Compliant Robotic Systems}, 
  year={2021},
  volume={},
  number={},
  pages={3292-3298},
  doi={10.1109/ICRA48506.2021.9561816}}

@Article{PagolicapacitiveSensor,
AUTHOR = {Pagoli, Amir and Chapelle, Frédéric and Corrales-Ramon, Juan-Antonio and Mezouar, Youcef and Lapusta, Yuri},
TITLE = {Large-Area and Low-Cost Force/Tactile Capacitive Sensor for Soft Robotic Applications},
JOURNAL = {Sensors},
VOLUME = {22},
YEAR = {2022},
NUMBER = {11},
ARTICLE-NUMBER = {4083},
URL = {https://www.mdpi.com/1424-8220/22/11/4083},
PubMedID = {35684706},
ISSN = {1424-8220},
DOI = {10.3390/s22114083}
}

@INPROCEEDINGS{HarberMagneticTactileSensor,
  author={Harber, Evan and Schindewolf, Evan and Webster-Wood, Vickie and Choset, Howie and Li, Lu},
  booktitle={2020 IEEE SENSORS}, 
  title={A Tunable Magnet-based Tactile Sensor Framework}, 
  year={2020},
  volume={},
  number={},
  pages={1-4},
  doi={10.1109/SENSORS47125.2020.9278634}}

@Article{MohammadiMagneticSensor,
AUTHOR = {Mohammadi, Alireza and Xu, Yangmengfei and Tan, Ying and Choong, Peter and Oetomo, Denny},
TITLE = {Magnetic-based Soft Tactile Sensors with Deformable Continuous Force Transfer Medium for Resolving Contact Locations in Robotic Grasping and Manipulation},
JOURNAL = {Sensors},
VOLUME = {19},
YEAR = {2019},
NUMBER = {22},
ARTICLE-NUMBER = {4925},
URL = {https://www.mdpi.com/1424-8220/19/22/4925},
PubMedID = {31726702},
ISSN = {1424-8220},
DOI = {10.3390/s19224925}
}

@misc{mi2024designvariablestiffnessquasidirect,
      title={Design of a Variable Stiffness Quasi-Direct Drive Cable-Actuated Tensegrity Robot}, 
      author={Jonathan Mi and Wenzhe Tong and Yilin Ma and Xiaonan Huang},
      year={2024},
      eprint={2409.05751},
      archivePrefix={arXiv},
      primaryClass={cs.RO},
      url={https://arxiv.org/abs/2409.05751}, 
}

@article{sipos2024gelslim,
  title={GelSlim 4.0: Focusing on touch and reproducibility},
  author={Sipos, Andrea and Bogert, William van den and Fazeli, Nima},
  journal={arXiv preprint arXiv:2409.19770},
  year={2024}
}

@inproceedings{xie2017aggregated,
  title={Aggregated residual transformations for deep neural networks},
  author={Xie, Saining and Girshick, Ross and Doll{\'a}r, Piotr and Tu, Zhuowen and He, Kaiming},
  booktitle={Proceedings of the IEEE conference on computer vision and pattern recognition},
  pages={1492--1500},
  year={2017}
}

@inproceedings{he2016deep,
  title={Deep residual learning for image recognition},
  author={He, Kaiming and Zhang, Xiangyu and Ren, Shaoqing and Sun, Jian},
  booktitle={Proceedings of the IEEE conference on computer vision and pattern recognition},
  pages={770--778},
  year={2016}
}

@inproceedings{stone2020walking,
  title={Walking on TacTip toes: A tactile sensing foot for walking robots},
  author={Stone, Elizabeth A and Lepora, Nathan F and Barton, David AW},
  booktitle={2020 IEEE/RSJ International Conference on Intelligent Robots and Systems (IROS)},
  pages={9869--9875},
  year={2020},
  organization={IEEE}
}

@article{song2024tactid,
  title={TacTID: high-performance visuo-tactile sensor-based terrain identification for legged robots},
  author={Song, Ziwu and Li, Chenchang and Quan, Zhentan and Mu, Shilong and Li, Xiaosa and Zhao, Ziyi and Jin, Wanxin and Wu, Chenye and Ding, Wenbo and Zhang, Xiao-Ping},
  journal={IEEE Sensors Journal},
  volume={24},
  number={16},
  pages={26487--26495},
  year={2024},
  publisher={IEEE}
}

@inproceedings{chorley2009development,
  title={Development of a tactile sensor based on biologically inspired edge encoding},
  author={Chorley, Craig and Melhuish, Chris and Pipe, Tony and Rossiter, Jonathan},
  booktitle={2009 International Conference on Advanced Robotics},
  pages={1--6},
  year={2009},
  organization={IEEE}
}

@article{ward2018tactip,
  title={The tactip family: Soft optical tactile sensors with 3d-printed biomimetic morphologies},
  author={Ward-Cherrier, Benjamin and Pestell, Nicholas and Cramphorn, Luke and Winstone, Benjamin and Giannaccini, Maria Elena and Rossiter, Jonathan and Lepora, Nathan F},
  journal={Soft robotics},
  volume={5},
  number={2},
  pages={216--227},
  year={2018},
  publisher={SAGE Publications Sage CA: Los Angeles, CA}
}


%% file: strings-abrv.bib
@STRING{C-IROS = {Proc. {IEEE}/{RSJ} Int. Conf. Intell. Robots and Syst.}}


%% file: strings-full.bib
@STRING{C-IROS = {Proceedings of the {IEEE}/{RSJ} International Conference on Intelligent Robots and Systems}}
